\documentclass{article}

     \usepackage[nonatbib, final]{neurips_2019}

\usepackage[utf8]{inputenc} % allow utf-8 input
\usepackage[T1]{fontenc}    % use 8-bit T1 fonts
\usepackage{hyperref}       % hyperlinks
\usepackage{url}            % simple URL typesetting
\usepackage{booktabs}       % professional-quality tables
\usepackage{amsfonts}       % blackboard math symbols
\usepackage{nicefrac}       % compact symbols for 1/2, etc.
\usepackage{microtype}      % microtypography
\usepackage{float}
\usepackage{graphicx}
\usepackage{subfig}

\usepackage{cite}

\title{Prediction of gaze direction using Convolutional Neural Networks for Autism diagnosis}

\author{
  \textbf{Dennis Núñez-Fernández$^1$, Franklin Porras-Barrientos$^1$} \\ \textbf{Macarena Vittet-Mondoñedo$^1$, Robert H. Gilman$^2$, Mirko Zimic$^1$}
  \\
  $^1$Laboratorio de Bioinformática y Biología Molecular, Universidad Peruana Cayetano Heredia, Peru \\ 
  $^2$Department of International Health, Johns Hopkins University, USA
  \\
  \texttt{\{dennis.nunez, franklin.barrientos.p, macarena.vittet.m\}@upch.pe} 
  \\
  \texttt{rgilman1@jhmi.edu, mirko.zimic@upch.pe} 
}

\begin{document}

\maketitle

\begin{abstract}
Autism is a developmental disorder that affects social interaction and communication of children. The gold standard diagnostic tools are very difficult to use and time consuming. However, diagnostic could be deduced from child gaze preferences by looking a video with social and abstract scenes. In this work, we propose an algorithm based on convolutional neural networks to predict gaze direction for a fast and effective autism diagnosis. Early results show that our algorithm achieves real-time response and robust high accuracy for prediction of gaze direction.
\end{abstract}

\section{Introduction}

Around 1 in 160 children worldwide is affected by Autism spectrum disorder (ASD). It generates a deficit social interaction \cite{1} and result in delay in cognitive development \cite{2}. Recent studies have shown that early intervention for children with ASD is effective in improving quality of life, every dollar spent on early intervention helps to save eight dollars in special education \cite{3, 4}. The reasons to the low utilization of the gold standard diagnostic tools are the duration of the tests and the extensive training for the technician \cite{5}, and developing countries have very few of them. Recent studies have shown strong evidence for utilizing gaze direction as an early biomarker of ASD \cite{6, 7, 8, 9}. Indeed, children with ASD show a preference for geometric scenes rather than social scenes \cite{8, 9}.

In recent years, several approaches to gaze direction recognition were proposed and some open source eye-tracking algorithms are currently available; however, these algorithms demand extensive calibration, several settings and training processes that are not appropriate for young children. For instance, most of current gaze direction systems for the ASD diagnosis need to be evaluated under controlled environments by using expensive devices that require holding the head to avoid undesirable movements \cite{8, 9}. None of such systems are appropriate in children due its restless behavior. 

In a more recent work \cite{10}, eye movements are used to asses ASD. Based on the gaze patterns and using K-means clustering and a support vector machine classifier, they are able to identify children with ASD with an accuracy of 88.51\%. Nonetheless, since this process involves a high-acuity eye tracker, it is not a scalable screening process. In \cite{11}, a CNN-based approach is used to predict gaze in a natural social interaction and assess ASD in children. They developed two CNNs, one for face detection and another for gaze prediction. Despite gaze direction is precisely predicted by the CNNs, wearing glasses involves equipment, which disturb child's attention.In addition, to date there are several popular open source tools for eye tracking. One of the most popular is \url{https://github.com/pupil-labs/pupil}, which provides an accurate tool for eye tracking, however, this system makes use of glasses. As explained above, external hardware is not suitable for children. Another popular tool for eye tracking is \url{http://www.pygaze.org}, however, calibration is difficult and eye region should be in a fixed position, which is difficult to set in children.

\section{Methodology}

\begin{figure}[H]
  \centering{\includegraphics[width=90mm]{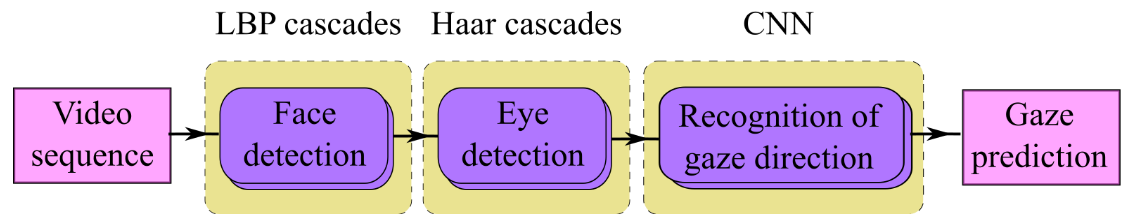}}
  \caption{Diagram for the proposed system}
  \label{diagram}
\end{figure}

Our proposed system recognizes gaze direction based on images obtained from a video sequence. Face and eye detection are performed by cascade classifiers using LBP and Haar features that were found with the Viola-Jones mechanism \cite{12}. Face detection uses LBP cascades over the whole image due its better performance, and eye detection employes Haar cascades into the facial region since they are much smaller than the full image. Later, we generate a single square image with both eye regions. Finally, the CNN classifies it into right, left or vague direction, see Fig.~\ref{diagram}.

The dataset was collected in our research facilities, the Laboratory of Bioinformatics and Molecular Biology, Universidad Peruana Cayetano Heredia, Peru. We enroled 30 adults between 22-35 years old, working in our laboratory. The videos were recorded under a controlled environment and using a standard web camera. The eye glace directons were three: right, left and vague. After frame extraction, we obtained a total of 420 images, which increased to 66,750 after data augmentation. 

The proposed CNN is a variation of the LeNet model \cite{13}, with 60K learnable parameters. The training of the proposed CNN has been carried out on the 80\% of the collected dataset (53,400 images) and testing on the remaining 20\% (13,350 images). The CNN input are 72x72 pixel binary images, following the architecture: C(5x5)-S(2x2)-C(5x5)-S(2x2)-FC(120)-FC(3), where C: Conv. layer, S: Sub sampling, FC: Full connection. We employed Caffe framework \cite{14}.

\section{Early Results}

For all shuffle testing on adult dataset, we obtain 96.01\% of accuracy. However, for a rigorous testing, we evaluated our model using a 5-fold cross-validation and using different groups of people who do not appear in the training dataset. For testing on the dataset using 3 classes and employing 5-fold cross-validation, we obtained an average accuracy of 89.54\%. Tests were conducted with people who do not appear in the training dataset.

\begin{figure}[H]
  \centering{\includegraphics[width=55mm]{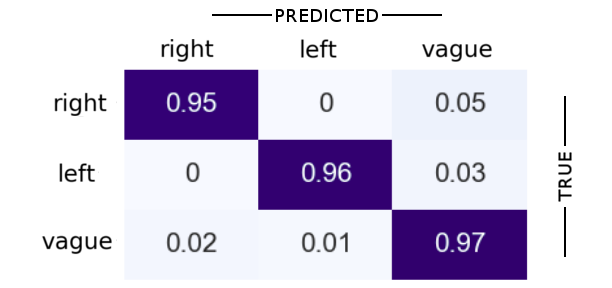}}
  \caption{Confusion matrix for three classes}
  \label{online_results}
\end{figure}

\section{Conclusions}

In this work we have presented the first results of a CNN-based methodology for eye glance prediction using a web camera with the aim to help in autism diagnosis. The system recognizes three gaze directions and works on a desktop PC. We show that our proposed method achieves a high classification accuracy of 96.01\% for testing. Furthermore, the system shows a real-time response of about 90 ms. The previous results demonstrate that the proposed system is a useful, fast, effective and accessible autism diagnosis tool.

\clearpage

\small

\bibliography{sample}{}
\bibliographystyle{plain}

\end{document}